\documentclass[runningheads]{llncs}
\usepackage[T1]{fontenc}
\usepackage{graphicx,verbatim}
\usepackage{cite}

\usepackage{enumitem,amssymb}
\newlist{todolist}{itemize}{2}
\setlist[todolist]{label=$\square$}
\usepackage{pifont}

\usepackage{booktabs}
\usepackage{multirow}

\usepackage{amsmath}

\usepackage{algorithm}
\usepackage{algpseudocode}
\usepackage{mathabx}
\usepackage[table]{xcolor}
\usepackage{hhline}
\usepackage{caption}
\usepackage{hyperref}

\newcolumntype{C}[1]{>{\centering\let\newline\\\arraybackslash\hspace{0pt}}m{#1}}

\algnewcommand{\IfThenElse}[3]{
  \State \quad \algorithmicif\ #1\ \algorithmicthen\ #2\ \algorithmicelse\ #3}

\newcommand{\confmethod}{\texttt{RUE}}
\newcommand{\transmethod}{\texttt{STF}}
\newcommand{\segmethod}{\texttt{STF-RUE}}

\newcommand{\confscp}{\texttt{PW-SCP}}
\newcommand{\adjconfscp}{\texttt{PW-SCP-ADJ}}
\newcommand{\confcrc}{\texttt{PW-CRC}}
\newcommand{\adjconfcrc}{\texttt{PW-CRC-ADJ}}

\definecolor{SP}{rgb}{0, 0, 0.66}
\newcommand{\sukanya}[1]{{#1}}

\newcommand{\yorick}[1]{{#1}} 

\usepackage[color=green!40]{todonotes}

\begin{document}
%

\title{Segmentation-Guided CT Synthesis with Pixel-Wise Conformal Uncertainty Bounds}

\titlerunning{Segmentation-Guided CT Synthesis with Conformal Uncertainty Bounds}
%

\author{David Vallmanya Poch\inst{1} \and
Yorick Estievenart\inst{1} \and
Elnura Zhalieva\inst{2} \and
Sukanya Patra\inst{1} \and
Mohammad Yaqub\inst{2} \and
Souhaib Ben Taieb\inst{2}}
\authorrunning{D. Vallmanya Poch et al.}
%
\institute{University of Mons, Belgium
\email{\{david.vallmanyapoch, yorick.estievenart, sukanya.patra\}@umons.ac.be}\\
\and
Mohamed bin Zayed University of Artificial Intelligence (MBZUAI), UAE
\email{\{elnura.zhalieva, mohammad.yaqub, souhaib.bentaieb\}@mbzuai.ac.ae}\\
}


\maketitle              

\begin{abstract}


\sukanya{Accurate dose calculations in proton therapy rely on high-quality CT images. While planning CTs (pCTs) serve as a reference for dosimetric planning, Cone Beam CT (CBCT) is used throughout Adaptive Radiotherapy (ART) to generate sCTs for improved dose calculations. Despite its lower cost and reduced radiation exposure advantages, CBCT suffers from severe artefacts and poor image quality, making it unsuitable for precise dosimetry. Deep learning-based CBCT-to-CT translation has emerged as a promising approach. Still, existing methods often introduce \textbf{anatomical inconsistencies} and lack \textbf{reliable uncertainty estimates}, limiting their clinical adoption. To bridge this gap, we propose \segmethod{}—a novel framework integrating two key components. First, \transmethod{}, a segmentation-guided CBCT-to-CT translation method that enhances anatomical consistency by leveraging segmentation priors extracted from pCTs. Second, \confmethod{}, a conformal prediction method that augments predicted CTs with pixel-wise conformal prediction intervals, providing clinicians with a robust reliability indicator. Comprehensive experiments using UNet++ and Fast-DDPM on two benchmark datasets demonstrate that \segmethod{} significantly improves translation accuracy, as measured by a novel soft-tissue-focused metric designed for precise dose computation. Additionally, \segmethod{} provides better-calibrated uncertainty sets for synthetic CT, reinforcing trust in synthetic CTs. By addressing both anatomical fidelity and uncertainty quantification, \segmethod{} marks a crucial step toward safer and more effective adaptive proton therapy. Code is available at \url{https://anonymous.4open.science/r/cbct2ct_translation-B2D9/}.}

\end{abstract}

\section{Introduction}

\begin{figure}[t]
    \centering
    \includegraphics[width=1\textwidth]{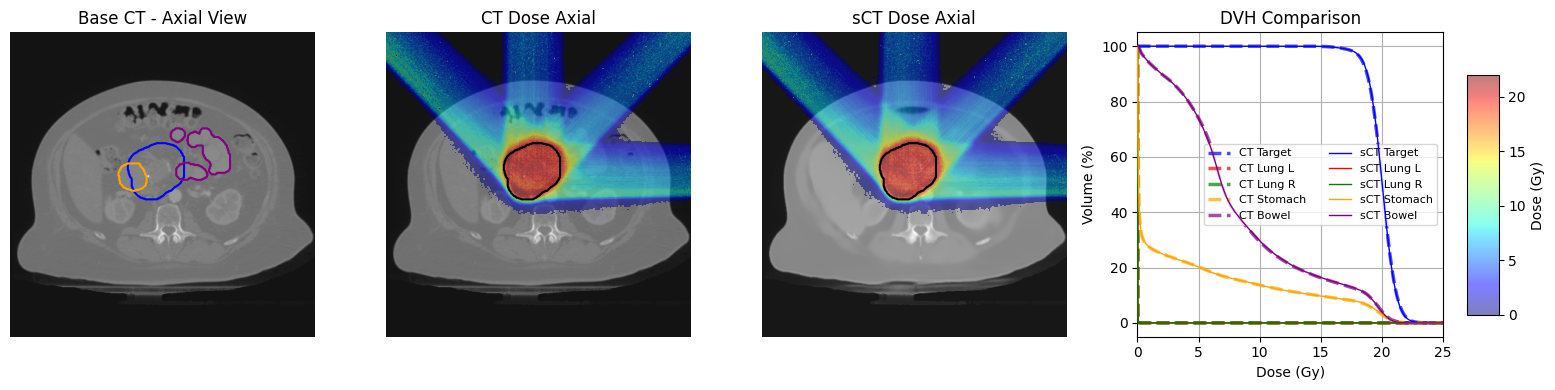}
    \caption{Comparison of dose distribution between CT and sCT from \segmethod{}, along with corresponding dose-volume histograms (DVH) from OpenTPS.}
    \label{fig:dose_maps}
\end{figure}

\sukanya{Proton radiotherapy \cite{yuan2019new, vanderwaeren2021clinical} is an advanced form of radiation therapy that uses high-energy proton beams to precisely target tumours while minimising damage to surrounding healthy tissues. Its advantages are particularly crucial for tumours located near critical structures and in pediatric oncology, where minimising radiation exposure is paramount \cite{Mohan2022-bk}. A pCT is acquired before treatment and serves as a high-quality reference for \textit{dosimetric planning}, enabling precise computation of radiation dose distribution to the tumour and surrounding healthy tissues. In contrast, a CBCT is used during treatment for patient positioning and anatomical monitoring. Despite its widespread use due to lower costs and reduced radiation exposure, CBCT suffers from image degradation caused by noise, scatter, and artefacts, making it unsuitable for dosimetric calculations. 


To address this challenge, CBCT-to-CT translation techniques have been developed to synthesize high-fidelity synthetic CT (sCT) that more accurately captures anatomical structures and tissue densities from CBCTs. Early approaches for CBCT-to-CT translation relied on intensity rescaling methods using lookup tables \cite{Kurz2015-sg} and histogram matching \cite{Arai2017-jx}, but these struggled to generalize across patients. Furthermore, Deformable Image Registration \cite{Peroni2012-fb, Veiga2015-jp, Veiga2016-lc} adapted pCTs to CBCT geometry but remained sensitive to registration errors and anatomical changes \cite{Niu2010-vn, Park2015-xh}. With the rise of deep learning, CNNs \cite{Kida2018-it}, UNet \cite{chen2021transunettransformersmakestrong, zhou2018unetnestedunetarchitecture}, GANs \cite{Isola2016-ji, Zhu2017-yp, Liang2019-sg} and diffusion models \cite{Lyu2022-xb, Fu2023-go, jiang2024fastddpmfastdenoisingdiffusion} have significantly improved translation accuracy. However, these deep learning approaches introduce two major limitations: (i) \textbf{Lack of anatomical consistency} in the generation of sCTs due to artificial hallucinations \cite{Angelopoulos2022-kj} and (ii) \textbf{Lack of reliable uncertainty estimates} for the sCTs that severely limits clinical trust in using the existing CBCT-to-CT translation approaches in critical applications such as proton radiotherapy. Some studies \cite{Rusanov2024-zi, Hemon2025-cw} have attempted to introduce uncertainty quantification via Monte Carlo dropout. However, they rely solely on asymptotic coverage guarantees, and if the underlying assumptions are violated, the uncertainty estimates may be unreliable, which is inadequate for high-stakes medical applications.}


\sukanya{To overcome the challenges, we propose \textbf{\underbar{S}egmentation-guided CBCT-to-CT \underbar{T}ranslation \underbar{F}ramework with 
\underbar{R}eliable \underbar{U}ncertainty \underbar{E}stimates} (\segmethod{}). We incorporate segmentation priors derived from pCTs to enhance structural fidelity in sCT generation, as illustrated by our dose distribution example in Figure~\ref{fig:dose_maps}. Specifically, we extract body and bone segmentations from pCTs using thresholding and grouping algorithms and then use them as additional inputs to guide the translation process. Furthermore, we augment the predicted CT with pixel-wise asymmetric conformal prediction intervals to provide reliable uncertainty estimates with finite-sample coverage guarantees \cite{Angelopoulos2022-kj, Horwitz2022-xs, belhasin2024principaluncertaintyquantificationspatial}. This allows clinicians to identify low-confidence regions and assess the reliability of generated sCTs. We implement \segmethod{} using two distinct architectures: UNet++ \cite{zhou2018unetnestedunetarchitecture}, a UNet-based model, and Fast-DDPM \cite{jiang2024fastddpmfastdenoisingdiffusion}, a diffusion-based approach. Our contributions are as follows: 


%

\begin{itemize}
    \item We introduce a segmentation-guided CBCT-to-CT translation method (\transmethod{}) that enhances anatomical consistency by incorporating segmentation priors extracted from pCTs.

    \item We propose a conformal prediction method (\confmethod{}) that augments predicted CTs with pixel-wise conformal prediction intervals, offering clinicians a robust measure of reliability.

    \item We assess our \segmethod{} framework for sCT synthesis on paired CBCT and CT scans from multiple patients. We evaluate tissue reconstruction in sCT using a novel metric, while empirical coverage and interval sizes assess the quality of uncertainty estimates.
    

    
\end{itemize}
}

\section{Segmentation-guided CBCT-to-CT Translation with Reliable Uncertainty Estimates}

We present our \segmethod{} framework, which integrates segmentation priors into CBCT-to-CT translation while providing reliable uncertainty estimation. First, in Section~\ref{sec:translation}, we discuss our segmentation-guided CBCT-to-CT translation method, \transmethod{}. Then, in Section~\ref{sec:confpred}, we introduce \confmethod{}, which augments \transmethod{} with robust uncertainty estimation.

\sukanya{We have access to a dataset \( \mathcal{D} = \{(X_i, \widetilde{Y}_i, Y_i)\}_{i=1}^{n} \), where \( X_i \in \mathcal{X} \) represents the CBCT, \( \widetilde{Y}_i \in \mathcal{Y} \) the corresponding pCT, and \( Y_i \in \mathcal{Y} \) the ground-truth CT. We define the input space as \( \mathcal{X} = [-1,1]^{D \times H \times W} \) and the output space as \( \mathcal{Y} = [-1,1]^{H \times W} \), assuming intensity normalization to \([-1,1]\). Our goal is to generate a sCT, denoted as \( \hat{Y} \), from \( X \) while leveraging segmentation priors extracted from \( \widetilde{Y} \). These priors are represented by bone and body segmentation masks, \( M^\text{bone}, M^\text{body} \in \mathcal{M} \), where \( \mathcal{M} = \{0,1\}^{H \times W} \). }

\subsection{Segmentation-Guided CBCT-to-CT Translation}
\label{sec:translation}

\sukanya{We propose a segmentation pipeline that leverages the direct relationship between intensity values and tissue radiodensity in pCTs to extract segmentation priors. Our approach builds on OpenTPS~\cite{wuyckens2023opentpsopensourcetreatment}. However, as OpenTPS struggles with detecting medium- and low-intensity bone regions, we augment it with a two-stage refinement algorithm that enhances the precision of anatomical feature extraction. First, we generate a binary mask for the body using intensity thresholding on \(\widetilde{Y}\), followed by morphological operations to remove small artefacts and fill holes, yielding a smooth body segmentation mask, \( M^\text{body} \). Next, we introduce a multi-threshold voxel classification technique for bone segmentation, categorising voxels into high-, medium-, and low-intensity regions. To ensure the continuity of bone structures, we dilate the high-intensity mask, effectively bridging gaps with medium-intensity regions. This results in an improved bone segmentation mask, \( M^\text{bone} \). Throughout the paper, we refer to the concatenation of the bone and body segmentation masks as the segmentation prior, \( M = M^\text{bone} \oplus M^\text{body} \), where \( \oplus \) denotes the concatenation operator.}

\begin{figure}[t]
    \centering
    \includegraphics[width=1\textwidth]{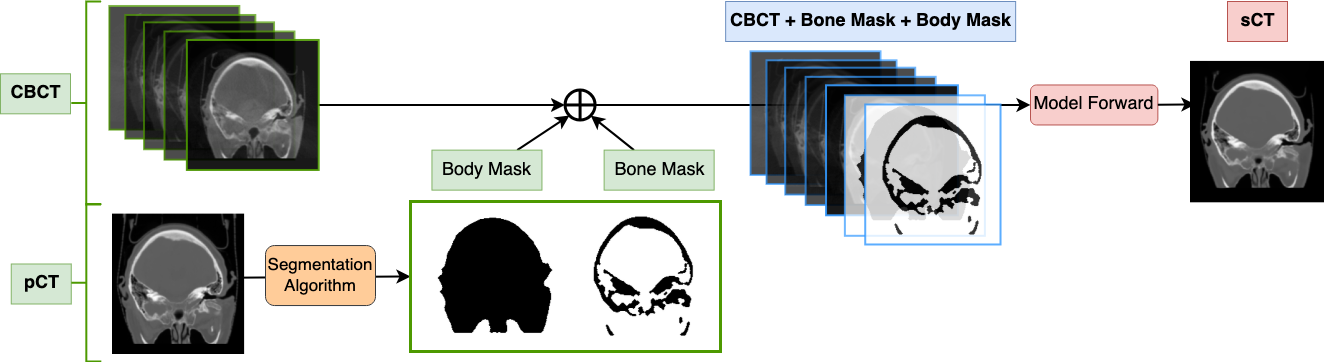}
    \caption{\sukanya{Overview of the proposed \transmethod{} Framework.}}
    \label{fig:pipeline_diagram}
\end{figure}



\sukanya{Translating CBCT to CT poses several challenges, such as registration inconsistencies, limited field-of-view (FOV), and low image quality. To address this, building on the segmentation priors, we propose to incorporate \(M\) as additional inputs for sCT generation. The design is guided by the assumption that while soft tissues deform over time during Adaptive Radiotherapy (ART), skeletal structures and body contours remain largely stable, allowing pCT-derived segmentation priors to be used for enhanced generation accuracy, thereby leading to improved dosimetry precision. By incorporating \(M\), \transmethod{} enforces anatomical consistency, delineates boundaries, improves soft tissue reconstruction in low-contrast regions, and corrects misalignment artefacts arising from registration discrepancies, as seen in Figure~\ref{fig:pipeline_diagram}.} \yorick{Since \transmethod{} uses pCTs, routinely available in practice, it can readily be adopted in existing clinical workflows.}

\subsection{Pixel-wise Prediction Intervals for sCT Reliability Assessment}
\label{sec:confpred}


Our \transmethod{} framework, discussed previously, enables the generation of an sCT image \( \hat{Y} \) for dosage calculation. However, it does not provide any information about the reliability of the prediction at each pixel. While the residual standard deviation can serve as an uncertainty estimate, it lacks rigorous coverage guarantees. Instead, we propose leveraging the Conformal Prediction (CP) framework \cite{Angelopoulos2021-st} to construct pixel-wise prediction intervals with reliable uncertainty estimates.

Formally, given an input $Z$ ($Z = X$, $Z = M$, or $Z = X \oplus M$) and an sCT $\hat{Y}$, we want to generate a pixel-wise prediction interval $\hat{C}(Z)_{i,j} = [\hat{Y}_{i,j} - \hat{l}(Z)_{i,j}, \hat{Y}_{i,j} + \hat{u}(Z)_{i,j}]$ for each pixel, where $i = 1, \dots, W$, $j = 1, \dots, H$, and $\hat{l}$ and $\hat{u}$ are lower and upper interval radii. Under the exchangeability assumption, CP intervals for a new test pair $(Z^{\text{test}}, Y^{\text{test}})$,  will satisfy the marginal coverage guarantee: $\mathbb{P}\big(Y^{\text{test}}_{i, j} \in \hat{C}(Z^{\text{test}})_{i, j}\big) \geq 1 - \alpha$ for $\alpha \in (0, 1)$ and all \(i\), \(j\). This guarantees that the true pixel value is contained within the predicted interval with probability at least $1 - \alpha$ \cite{Barber2022-wh}. The widely used split CP algorithm (SCP) \cite{Angelopoulos2021-st} relies on a calibration set \( \mathcal{D}_C \subset \mathcal{D} \) of size \( n_C \). Given a score function \( s: [-1, 1] \times [-1, 1] \to \mathbb{R} \), a conformity score is assigned to each calibration sample as \( S^k_{i, j} = s(X^k_{i, j}, Y^k_{i, j}) \) for \( k = 1, \dots, n_C \). A conformal quantile is computed as \( \hat{q}_{i, j} = S^{\lceil (1 - \alpha) (n_C + 1) \rceil}_{i, j} \), and the resulting CP interval is given by \( \hat{C}(Z^{\text{test}})_{i, j} = \left\{ Y_{i, j} \in [-1, 1] \mid s(Z^{\text{test}}_{i, j}, Y_{i, j}) \leq \hat{q}_{i, j} \right\} \). If the absolute error is used as the score function, i.e., \( s(Z^k_{i, j}, Y^k_{i, j}) = \lvert \hat{Y}^k_{i, j} - Y^k_{i, j} \rvert \), the prediction interval is symmetric, satisfying \( \hat{l}(Z^{\text{test}})_{i, j} = \hat{u}(Z^{\text{test}})_{i, j} = \hat{q}_{i, j} \). We refer to this approach as \confscp, where \texttt{PW} stands for pixel-wise.

Adaptive prediction intervals that adjust based on the input \( Z^{\text{test}} \) are often desirable, but \confscp{} lacks this flexibility. Instead of constructing prediction intervals centered at pixel-wise values, we first compute heuristic prediction intervals by extracting pixel-wise lower (\(\tilde{l}\)) and upper (\(\tilde{u}\)) bounds from sCT samples generated by a diffusion model. We then refine these intervals using conformal risk control (CRC) \cite{Angelopoulos2022-am}, adjusting the bounds as follows: \(\hat{l}(Z^{\text{test}})_{i, j} = \tilde{l}(Z^{\text{test}})_{i, j} - \hat{\lambda}\) and \(\hat{u}(Z^{\text{test}})_{i, j} = \tilde{u}(Z^{\text{test}})_{i, j} + \hat{\lambda}\), where \(\hat{\lambda} \in \mathbb{R}^+\) is determined by CRC. CRC ensures that the risk function remains below a predefined threshold \(\alpha\) by solving \(\hat{\lambda} = \inf \left\{ \lambda : \frac{n_C}{n_C + 1} \hat{R}_{n_C}(\lambda) + \frac{B}{n_C + 1} \leq \alpha \right\}\), where \(\hat{R}_{n_C}(\lambda)\) is the mean risk function computed on the calibration set, and \(B\) is an upper bound on the risk. We use the miscoverage loss as the risk function to enforce marginal coverage guarantees. Our approach, referred to as \confcrc{}, differs from that of \cite{Horwitz2022-xs} in three key ways: (i) it provides precise coverage guarantees rather than risk-controlling prediction sets \cite{Bates2021-gj}, (ii) it leverages CRC to adjust the interval bounds, and (iii) it employs additive updates instead of multiplicative ones, ensuring stability when interval radii are close to zero.

Finally, another challenge arises from the fact that the exchangeability assumption does not hold across all slices extracted from a CBCT due to spatial correlations and shared anatomical structures. To address this, we assume exchangeability at the patient level rather than within individual slices, making the CP guarantee primarily dependent on the number of patients, \( P \). A similar assumption was considered in \cite{Lu2023-qf} in the context of federated learning. For \confscp{}, we adjust the quantile computation to account for \( P \) and the average number of samples per patient, \( n_P = \frac{n_C}{P} \), modifying the quantile estimate as \( \hat{q}_{i, j} = S^{\lceil (1 - \alpha) (n_C + n_P) \rceil}_{i, j} \), and denote this adjusted version as \adjconfscp{}. Similarly, for \confcrc{}, we adjust the computation by setting \( \hat{\lambda} = \inf \left\{ \lambda : \frac{n_C}{n_C + n_P} \hat{R}_{n_C}(\lambda) + \frac{B \, n_P}{n_C + n_P} \leq \alpha \right\} \), and denote this method as \adjconfcrc{}.

\section{Experiments}

\noindent \textbf{Experimental setup}. \sukanya{We conduct our experiments using two publicly available datasets: (i) SynthRAD \cite{thummerer2023synthrad}, which includes 180 CBCT/CT pairs with patient outline masks for brain and pelvis regions, referred hereafter as S-Brain and S-Pelvis, respectively, and (ii) Pancreatic-CT-CBCT-SEG (Pancreas) \cite{hong2021breathhold}, comprising 40 patients with locally advanced pancreatic cancer, each with one CT and two CBCT scans acquired during deep inspiration breath-hold. All images originally had intensity ranges in Hounsfield Units (HUs), which are min-max normalized to \([-1,1]\) and center-cropped to \(416\times416\) for Pancreas and S-Pelvis, and \(288\times288\) for S-Brain. Due to computational constraints, the Pancreas Fast-DDPM model uses a resolution of \(128\times128\). Affine transformations are applied as data augmentation. The data is split into 70\% for training and 30\% for testing. For CP, the test set is further divided into 20\% for calibration and 10\% for final testing. We evaluate two baseline models: UNet++ \cite{zhou2018unetnestedunetarchitecture}, and Fast-DDPM \cite{jiang2024fastddpmfastdenoisingdiffusion}. 

A key challenge in evaluating our method arises from the absence of pCTs in the publicly available datasets. To address this, we use the provided CTs \(Y\) to extract segmentation priors \(M\) for \segmethod{}. Note that while we use ground-truth CTs to derive segmentation priors, the CTs are real-valued in the range $[-1,1]$, whereas the segmentation priors are binary, discarding substantial information. This inherently mitigates the risk of information leakage.
To evaluate the efficacy of our method, we assess its performance under three input settings: (1) CBCT-only (CBCT), where input is only CBCT; (2) segmentation priors-only (SEG), where only segmentation priors are provided; and (3) CBCT with segmentation priors (C+SEG), where both inputs are used. We further analyze robustness by adding noise perturbations to the CTs before extracting segmentation priors.  

UNet++ is implemented with a ResNeXt-101 backbone as validated in the SynthRAD2023 challenge report \footnote{For further details refer to the \href{https://arxiv.org/abs/2403.08447v3}{SynthRAD2023 challenge report}}. Fast-DDPM, is conditioned on the three baseline input settings and trained over \(40\) uniformly sampled time steps with default parameters from \cite{jiang2024fastddpmfastdenoisingdiffusion}.} \yorick{Within the CP framework, we apply \confscp{} using UNet++ and perform \confcrc{} sampling with Fast-DDPM. The miscoverage level is set to $\alpha = 0.1$. For the SynthRAD dataset, $P = 39$ (brain, pelvis) and $P = 10$ for Pancreas, while $n_P$ is \(73\) (pelvis), \(204\) (brain), and \(80\) (pancreas).}

\noindent \textbf{Evaluation metrics}. \sukanya{To evaluate tissue reconstruction in sCT, we report SoftMAE, a new metric we propose, along with the standard MAE. To calculate SoftMAE, we first extract body mask $M^{\text{body}}$ and bone mask $M^{\text{bone}}$ from ground-truth CT as described in Section \ref{sec:translation}. Similarly, we derive the body and bone masks, $\hat{M}^{\text{body}}$ and $\hat{M}^{\text{bone}}$, from sCT. Next, we define a mask \(M^{\text{soft}} = ( M^{\text{body}} \cap \hat{M}^{\text{body}}) \setminus (M^{\text{bone}} \cup \hat{M}^{\text{bone}})\) that highlights soft tissues while excluding high-contrast bone regions, minimising registration related errors. Then, given sCT and ground-truth CT slices \(\hat{Y}, Y\), $\text{SoftMAE} = \sum_{i,j}^{H,W} W_{i,j} |\hat{Y}_{i,j} - Y_{i,j}|$, where $W = M^{\text{soft}}/\sum_{i,j}^{H,W} M_{i,j}^{\text{soft}}$. Additionally, given two masks \( p \) and \( q \), we compute the Dice coefficient as \( \text{Dice}(p, q) = 2\frac{ |p \cap q|}{|p| + |q|} \)for body and bone masks as a complementary measure of segmentation accuracy. Specifically, $\text{DiceBody} = \text{Dice}(M^{\text{body}}, \hat{M}^{\text{body}})$ and $\text{DiceBone} = \text{Dice}(M^{\text{bone}}, \hat{M}^{\text{bone}})$.
\yorick{We compute pixel-wise marginal empirical coverage (i.e., proportion of ground-truth values within predicted intervals) and prediction interval size (i.e., mean pixel-wise distance between bounds across test samples). Since marginal coverage alone may be misleading (e.g., when intervals are excessively wide), we also report pixel-stratified coverage by grouping pixel values into $G$ groups and measuring the error at the specified coverage level within each group, i.e., as \((\sum_{g \in G} |(1 - \alpha) - \text{Cov}(g)|)/|G|\). Metrics are computed within masked regions.}}

\begin{table}[t]
    \centering
    \caption{Comparison with baselines. MAE and SoftMAE are shown in HUs.}
    \aboverulesep = 0pt
    \belowrulesep = 0pt
    \setlength{\tabcolsep}{1.5pt}
    \renewcommand{\arraystretch}{0.85}
    \fontsize{8pt}{12}\selectfont
    \begin{tabular}{|>{\centering\arraybackslash}p{0.4cm}|l|l|c|c|cc|>{\centering\arraybackslash}p{0.4cm}|cc|cc|cc|}
        \toprule
         \multirow{3}{*}{} & \multirow{3}{*}{\textbf{Dataset}} & \multirow{3}{*}{\textbf{Conf.}} & \multicolumn{4}{c|}{\textbf{CBCT-to-CT translation}} &  & \multicolumn{6}{c|}{\textbf{Uncertainty quantification}}\\
         \cmidrule{4-14}
         &  &  & \multirow{2}{*}{\textbf{MAE}$\downarrow$} & \multicolumn{1}{c|}{\textbf{Soft}} & \multicolumn{2}{c|}{\textbf{Dice}$\uparrow$} & \multirow{2}{*}{} & \multicolumn{2}{c|}{\textbf{M-Cov. $\uparrow$}} & \multicolumn{2}{c|}{\textbf{P-Cov. $\downarrow$}} & \multicolumn{2}{c|}{\textbf{Int. Size $\downarrow$}} \\
        & & & & \multicolumn{1}{c|}{\textbf{MAE}$\downarrow$} & \textbf{Body} & \textbf{Bone} & & \textbf{Base} & \textbf{Adj.} & \textbf{Base} & \textbf{Adj.} & \textbf{Base} & \textbf{Adj.} \\
        \midrule
        \multirow{9}{*}{\rotatebox[origin=c]{90}{\textsc{UNet++}}} & \multirow{3}{*}{S-Pelvis}
        & CBCT & 58.29 & 23.01 & 0.97 & 0.86 & \multirow{9}{*}{\rotatebox[origin=c]{90}{\confscp{}}} & 0.86 & 0.88 & 0.31 & 0.28 & 0.12 & 0.16 \\
        & & SEG & 38.74 & 27.76 & 0.99 & 0.99 & & 0.89 & 0.91 & 0.29 & 0.27 & 0.08 & 0.10 \\
         & & \cellcolor{gray!10}C+SEG & \cellcolor{gray!10} 29.46 & \cellcolor{gray!10} 20.34 & \cellcolor{gray!10} 1.00 & \cellcolor{gray!10} 0.98 & & \cellcolor{gray!10} 0.87 & \cellcolor{gray!10} 0.90 & \cellcolor{gray!10} 0.28 & \cellcolor{gray!10} 0.25 & \cellcolor{gray!10} 0.06 & \cellcolor{gray!10} 0.07 \\
        \cmidrule{2-7}\cmidrule{9-14}
        & \multirow{3}{*}{S-Brain}
        & CBCT & 58.39 & 20.25 & 0.97 & 0.96 & & 0.85 & 0.94 & 0.54 & 0.41 & 0.37 & 0.80 \\
        & & SEG & 59.95 & 21.63 & 1.00 & 1.00 & & 0.89 & 0.92 & 0.31 & 0.26 & 0.14 & 0.18 \\
        & & \cellcolor{gray!10}C+SEG & \cellcolor{gray!10} 40.00 & \cellcolor{gray!10} 17.41 & \cellcolor{gray!10} 1.00 & \cellcolor{gray!10} 1.00 & & \cellcolor{gray!10} 0.88 & \cellcolor{gray!10} 0.90 & \cellcolor{gray!10} 0.24 & \cellcolor{gray!10} 0.21 & \cellcolor{gray!10} 0.09 & \cellcolor{gray!10} 0.11 \\
        \cmidrule{2-7}\cmidrule{9-14}
        & \multirow{3}{*}{Pancreas}
        & CBCT & 135.89 & 60.15 & 0.92 & 0.70 & & 0.85 & 0.94 & 0.54 & 0.41 & 0.37 & 0.80 \\
        & & SEG & 77.47 & 79.98 & 0.99 & 0.98 & & 0.85 & 0.98 & 0.48 & 0.39 & 0.51 & 0.87 \\
        & & \cellcolor{gray!10}C+SEG & \cellcolor{gray!10} 59.04 & \cellcolor{gray!10} 59.85 & \cellcolor{gray!10} 0.99 & \cellcolor{gray!10} 0.98 & & \cellcolor{gray!10} 0.87 & \cellcolor{gray!10} 0.99 & \cellcolor{gray!10} 0.51 & \cellcolor{gray!10} 0.40 & \cellcolor{gray!10} 0.43 & \cellcolor{gray!10} 0.85 \\
        \midrule
        \multirow{9}{*}{\rotatebox[origin=c]{90}{\textsc{Fast-DDPM}}} & \multirow{3}{*}{S-Pelvis}
        & CBCT & 77.05 & 44.76 & 0.97 & 0.87 & \multirow{9}{*}{\rotatebox[origin=c]{90}{\confcrc{}}} & 0.81 & 0.86 & 0.49 & 0.46 & 0.16 & 0.18 \\
        & & SEG & 55.87 & 42.84 & 1.00 & 0.98 & & 0.92 & 0.94 & 0.24 & 0.23 & 0.13 & 0.14 \\
        & & \cellcolor{gray!10}C+SEG & \cellcolor{gray!10} 52.13 & \cellcolor{gray!10} 39.16 & \cellcolor{gray!10} 1.00 & \cellcolor{gray!10} 0.95 & & \cellcolor{gray!10} 0.89 & \cellcolor{gray!10} 0.91 & \cellcolor{gray!10} 0.28 & \cellcolor{gray!10} 0.27 & \cellcolor{gray!10} 0.10 & \cellcolor{gray!10} 0.11 \\
        \cmidrule{2-7}\cmidrule{9-14}
        & \multirow{3}{*}{S-Brain}
        & CBCT & 77.20 & 39.55 & 0.97 & 0.96 & & 0.57 & 0.60 & 0.54 & 0.51 & 0.15 & 0.18 \\
        & & SEG & 78.93 & 36.43 & 1.00 & 0.99 & & 0.90 & 0.92 & 0.22 & 0.18 & 0.15 & 0.18 \\
        & & \cellcolor{gray!10}C+SEG & \cellcolor{gray!10} 53.60 & \cellcolor{gray!10} 26.42 & \cellcolor{gray!10} 0.99 & \cellcolor{gray!10} 0.99 & & \cellcolor{gray!10} 0.86 & \cellcolor{gray!10} 0.89 & \cellcolor{gray!10} 0.32 & \cellcolor{gray!10} 0.30 & \cellcolor{gray!10} 0.13 & \cellcolor{gray!10} 0.15 \\
        \cmidrule{2-7}\cmidrule{9-14}
        & \multirow{3}{*}{Pancreas}
        & CBCT & 160.28 & 95.05 & 0.94 & 0.76 & & 0.79 & 0.92 & 0.41 & 0.10 & 0.41 & 0.92 \\
        & & SEG & 99.06 & 91.43 & 0.99 & 0.97 & & 0.90 & 0.99 & 0.17 & 0.09 & 0.17 & 0.80 \\
        & & \cellcolor{gray!10}C+SEG & \cellcolor{gray!10} 84.60 & \cellcolor{gray!10} 81.20 & \cellcolor{gray!10} 1.00 & \cellcolor{gray!10} 0.94 & & \cellcolor{gray!10} 0.85 & \cellcolor{gray!10} 0.99 & \cellcolor{gray!10} 0.23 & \cellcolor{gray!10} 0.09 & \cellcolor{gray!10} 0.23 & \cellcolor{gray!10} 0.82 \\
        \bottomrule
    \end{tabular}
    \label{tab:test_results}
\end{table}

\subsection{Results}

\begin{figure}[t]
    \centering
    \includegraphics[height=3cm, width=0.9\textwidth]{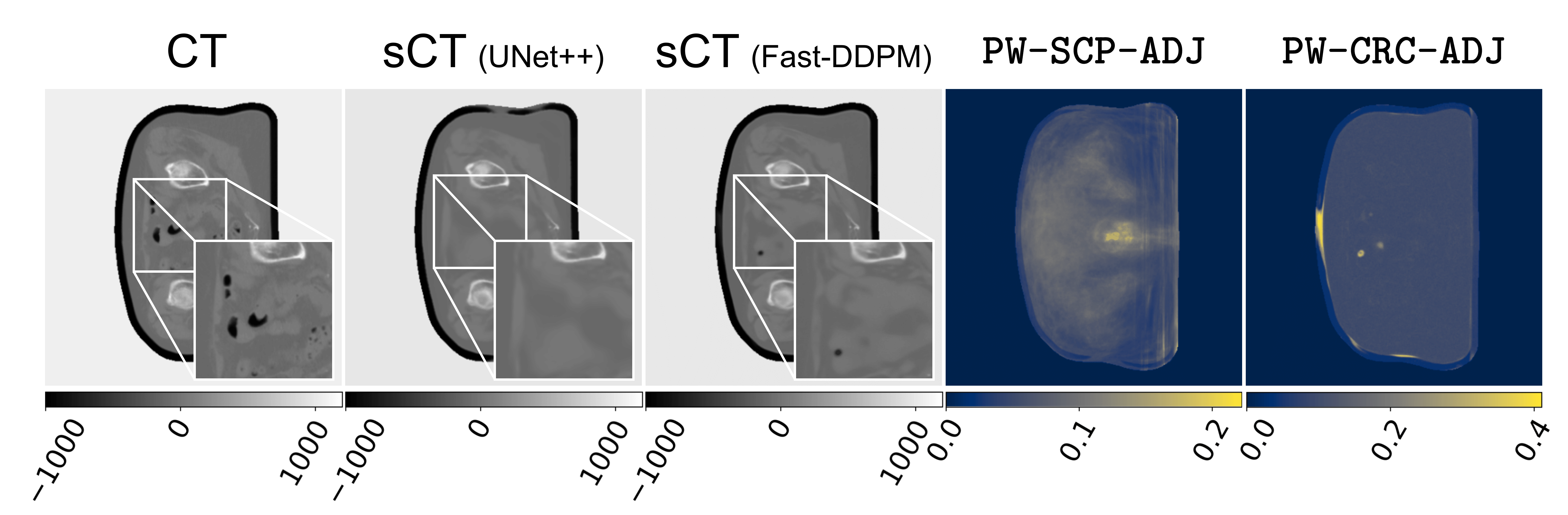}
    \caption{\yorick{Examples of sCTs from UNet++ and the diffusion model using C+SEG, with uncertainty maps from \adjconfscp{} and \adjconfcrc{}. CTs displayed in HU.}}
    \label{fig:uncertainty_maps}
\end{figure}

\begin{figure}[t]
    \centering
    \includegraphics[width=1\textwidth]{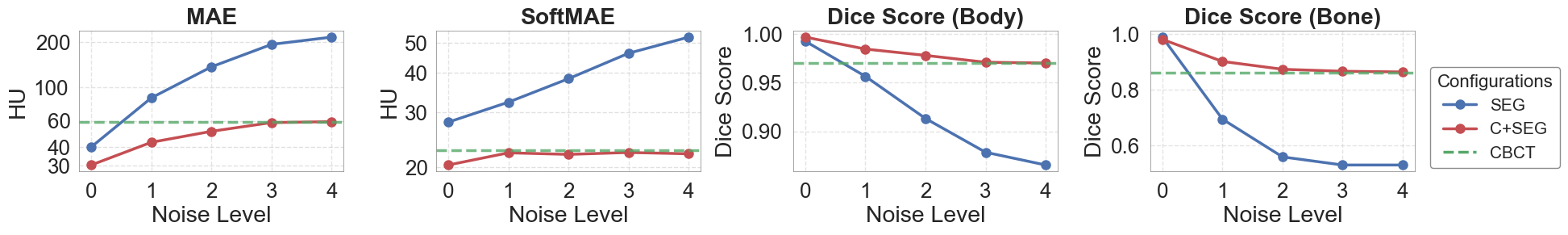}
    \caption{Effect of segmentation noise on model performance on SEG and C+SEG UNet++ models for the SynthRAD pelvis dataset.}
    \label{fig:affine_noise_levels}
\end{figure} 


\sukanya{\noindent \textbf{CBCT-to-CT translation}.} Results in the left side of Table \ref{tab:test_results} show that across all datasets and architectures, C+SEG outperforms SEG and CBCT-only models. In S-Pelvis, UNet++ with C+SEG achieves the lowest MAE ($29.46$) and SoftMAE ($20.34$). UNet++ with CBCT-only has a higher MAE but a lower SoftMAE than the SEG model, likely due to MAE’s sensitivity to high-contrast regions like bones, which can amplify CBCT registration errors. SoftMAE, focused on soft tissues, is less affected by these errors and is a more reliable metric for evaluation. In the Pancreas dataset, UNet++ with CBCT-only struggles due to limited FOV but captures soft tissue details, reflected in its lower SoftMAE than SEG. Both UNet++ and Fast-DDPM with C+SEG yield the lowest errors. However, on the same dataset with the same input setting, Fast-DDPM has a higher MAE but produces sharper, more detailed textures, as shown in the left panel of Figure~\ref{fig:uncertainty_maps}. Finally, SEG and C+SEG models consistently achieve higher Dice scores than CBCT-only, showing the importance of segmentation priors. 

\noindent \textbf{Impact of Perturbation of pCTs}. 
To evaluate the robustness of our method, we introduce noise at five levels, from no perturbation (0) to severe distortions (4), with up to $\pm15^\circ$ rotation, $\pm15\%$ translation, and $\pm15\%$ scaling. In Figure ~\ref{fig:affine_noise_levels}, we compare SEG and C+SEG input settings and observe that both degrade with noise, but SEG is more sensitive, showing a steeper error increase. C+SEG stabilizes, especially in SoftMAE, highlighting CBCT’s role in preserving soft tissue accuracy despite the segmentation noise.

\yorick{\noindent \textbf{Uncertainty Estimates}. The right side of Table \ref{tab:test_results} presents results for \confscp{} (using UNet++) and our \confcrc{} (using Fast-DDPM) across all test sets. The C+SEG model achieves the smallest interval sizes, except for the Pancreas dataset, likely due to the limited number of patients. The methods (\adjconfscp{} and \adjconfcrc{}) ensure marginal coverage and reduce pixel-stratified coverage error, albeit at the cost of wider prediction intervals. The CBCT-only model proves less reliable, with marginal coverage dropping even after adjustment, particularly in S-Brain. As shown in the right panel of Figure \ref{fig:uncertainty_maps}, \adjconfscp{} produces uniform uncertainty maps, defined as $\log(| \hat{C}(Z) | + 1)$, whereas \adjconfcrc{} highlights artefacts and hallucinated regions.}


\section{Conclusion}

CBCT is integral to ART workflows for its low cost and reduced radiation exposure, but its limitations in accurate dosimetry remain a challenge. Our segmentation-guided CBCT-to-CT translation method enhances anatomical consistency in sCT and provides robust pixel-wise CP intervals. Figure~\ref{fig:dose_maps} shows its impact on dose calculations. Evaluations with UNet++ and Fast-DDPM confirm the superiority of segmentation priors over CBCT-only inputs, particularly in soft-tissue translation accuracy, as measured by the SoftMAE metric. Additionally, \segmethod{} demonstrates robustness to variations in available pCTs, reinforcing its clinical viability. By bridging imaging limitations and precise treatment planning, our approach enhances dosimetric accuracy and equips clinicians with reliable uncertainty maps, enabling safer and more adaptive proton therapy.


\clearpage

\section*{Acknowledgement}

This work is supported by the research project Federated Learning and Augmented Reality for Advanced Control Centers. We sincerely thank Geoffroy HERBIN from IBA for his valuable discussions and insightful feedback, which greatly contributed to our understanding of the medical use case and dataset. His expertise helped us better grasp real-world challenges and refine our approach to address them effectively.

\bibliography{biblio_wo_links}
\bibliographystyle{splncs04}

\end{document}